\documentclass[letterpaper, 10 pt, conference]{ieeeconf}  % Comment this line out if you need a4paper
\IEEEoverridecommandlockouts
\usepackage{cite}
\usepackage{amsmath,amssymb,amsfonts}
\usepackage{algorithmic}
\usepackage{graphicx}
\usepackage{textcomp}
\usepackage[dvipsnames]{xcolor}

\usepackage[none]{hyphenat}
\usepackage{graphics} % for pdf, bitmapped graphics files
\graphicspath{{figures/}}
\usepackage[caption=false, font=normalsize, labelfont=sf, textfont=sf]{subfig}
\usepackage{physics}
\usepackage{textcomp}
\usepackage{mathtools}
\usepackage{blindtext}
\usepackage{tabularx}
\usepackage{multirow}
\usepackage{tabularx,booktabs}
\usepackage{lipsum}
\usepackage{cases}
\usepackage[short]{optidef}
\usepackage{url}
\usepackage[normalem]{ulem}
\usepackage[hidelinks]{hyperref}
\usepackage[linesnumbered,ruled,vlined]{algorithm2e}
\usepackage{soul}
\usepackage{comment}

\newcommand{\cmmnt}[1]{}

\usepackage{tikz}
\newcommand\copyrighttext{%
  \footnotesize \textcopyright 2025 IEEE. Personal use of this material is permitted.
  Permission from IEEE must be obtained for all other uses, in any current or future
  media, including reprinting/republishing this material for advertising or promotional
  purposes, creating new collective works, for resale or redistribution to servers or
  lists, or reuse of any copyrighted component of this work in other works.}
\newcommand\copyrightnotice{%
\begin{tikzpicture}[remember picture,overlay]
\node[anchor=north,yshift=-10pt] at (current page.north) 
  {\fbox{\parbox{\dimexpr\textwidth-\fboxsep-\fboxrule\relax}{\copyrighttext}}};
\end{tikzpicture}%
}

\def\BibTeX{{\rm B\kern-.05em{\sc i\kern-.025em b}\kern-.08em
    T\kern-.1667em\lower.7ex\hbox{E}\kern-.125emX}}
    
\begin{document}

\title{\LARGE \bf Model-Structured Neural Networks to Control the Steering Dynamics of Autonomous Race Cars}

\author{Mattia Piccinini$^{1*}$, Aniello Mungiello$^{2*}$, Georg Jank$^{3}$, Gastone Pietro Rosati Papini$^{4}$,\\Francesco Biral$^{4}$, Johannes Betz$^{1}$
\thanks{This work was partly funded by the Ministero Università e Ricerca (MUR), Italy, through the PNRR project Centro Nazionale per la Mobilità Sostenibile (CNMS) CUP E63C22000930007.}
\thanks{$^{*}$These authors contributed equally to this work.}
\thanks{$^{1}$Professorship of Autonomous Vehicle Systems, Technical University of Munich, 85748 Garching, Germany; Munich Institute of Robotics and Machine Intelligence (MIRMI), {\tt\small name.surname@tum.de}}
\thanks{$^{2}$Department of Information Technology and Electrical Engineering (DIETI), University of Naples Federico II, Naples 80125, Italy, {\tt\small aniello.mungiello@unina.it}}
\thanks{$^{3}$Chair of Automatic Control, Department of Mechanical Engineering, TUM School of Engineering and
Design, Technical University of Munich, 85748 Garching, Germany, {\tt\small georg.jank@tum.de}}
\thanks{$^{4}$Department of Industrial Engineering, University of Trento, 38123 Trento, Italy, {\tt\small name.surname@unitn.it}}
}

\copyrightnotice
\maketitle

\begin{abstract}
Autonomous racing has gained increasing attention in recent years, as a safe environment to accelerate the development of motion planning and control methods for autonomous driving. 
%This has led to the development of advanced algorithms in vehicle perception, motion planning, and control. 
Deep learning models, predominantly based on neural networks (NNs), have demonstrated significant potential in modeling the vehicle dynamics and in performing various tasks in autonomous driving. However, their black-box nature is critical in the context of autonomous racing, where safety and robustness demand a thorough understanding of the decision-making algorithms. To address this challenge, this paper proposes MS-NN-steer, a new Model-Structured Neural Network for vehicle steering control, integrating the prior knowledge of the nonlinear vehicle dynamics into the neural architecture. The proposed controller is validated using real-world data from the Abu Dhabi Autonomous Racing League (A2RL) competition, with full-scale autonomous race cars. In comparison with general-purpose NNs, MS-NN-steer is shown to achieve better accuracy and generalization with small training datasets, while being less sensitive to the weights' initialization. 
%while preserving the physical interpretability of its parameters. 
%We demonstrate that this physics-informed design leads to a more interpretable neural model, requiring fewer trainable parameters while achieving improved accuracy and generalization capabilities.
Also, MS-NN-steer outperforms the steering controller used by the A2RL winning team.
%, which is based on a combination of kinematic and empirical models. 
Our implementation is available open-source in the following repository \url{https://github.com/tonegas/nnodely-applications/tree/main/vehicle/control_steer_dynamics_A2RL}. 
\end{abstract}

\section{Introduction}
Over the past decade, autonomous driving has seen a steady growth, reflected in a surge of research addressing the many tasks required for vehicle autonomy. However, much of this work remains centered on standard urban scenarios, which, while important, often fail to capture the challenges of safety-critical conditions where vehicles operate near their dynamic limits.

To bridge this gap, autonomous racing has emerged as a valuable benchmark for pushing the boundaries of perception, planning, and control algorithms \cite{betz2022autonomous,Pagot2020,Piccinini2024_ggv}. At the limits of handling, the vehicle dynamics become highly nonlinear and sensitive to factors such as tire characteristics, suspensions, and aerodynamics, many of which vary with wear, temperature, and weather. This complexity often makes physics-based modeling methods impractical \cite{10384847}, prompting a growing interest in data-driven techniques, especially deep learning \cite{kuutti2020survey}. However, data-driven approaches still face challenges in generalizing to unseen conditions \cite{piccinini2023predictive}, especially with limited training data. Let us now critically analyze the related work, to later highlight the contributions of this paper.

% as they are unable to know the key vehicle parameters due to varying during operation and over the vehicle's lifetime. The identification of numerous interdependent components, such as tires, suspension systems, powertrain, braking, steering, and aerodynamic characteristics, can be both time-consuming and resource-intensive \cite{10384847}. Consequently, a growing number of studies address this problem using approaches that are not physics-based, such as neural networks \cite{kuutti2020survey}. 

\begin{figure}[]
    \centering
    \includegraphics[width=\columnwidth]{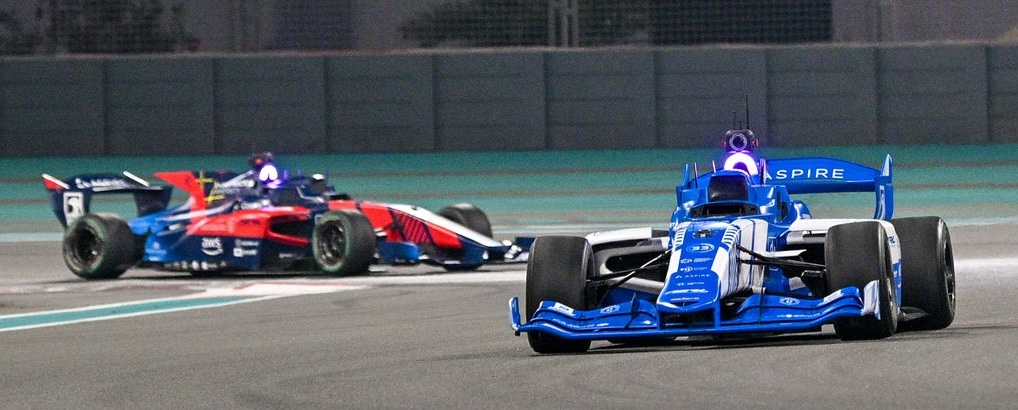}
    \caption{Autonomous race car of the Technical University of Munich (TUM) at the 2024 Abu Dhabi Autonomous Racing League (A2RL) competition, on the Yas Marina circuit.}
    \label{fig:A2RL}
\end{figure}

\subsection{Related Work}
Deep neural networks (NNs) have demonstrated significant potential in vehicle dynamics modeling \cite{hermansdorfer2021end, nie2022deep} and control \cite{kuutti2020survey,piccinini2023predictive,Spielberg2019,10388422,Weiss2020,Gottschalk2024}.
However, large-scale NNs typically require extensive training datasets to perform reliably in unseen scenarios \cite{kuutti2020survey,piccinini2023predictive}. Indeed, the general-purpose internal structure of NNs enables them to approximate a wide range of systems' behaviors, but at the price of requiring many trainable parameters and data.  
% This limited generalization is primarily due to the general-purpose internal structure of NNs. They are highly flexible and can approximate a wide range of system behaviors. However, this flexibility comes at the cost of a substantial number of trainable parameters and large amounts of data to ensure robust performance.\\
Moreover, deep NNs are often perceived as black boxes \cite{piccinini2023predictive}, making it difficult to assign physical meaning to their parameters and outputs. 
%posing significant challenges in terms of interpretability, explainability, and debuggability. 
This is particularly critical in the context of autonomous driving, where safety and robustness demand a proper understanding of the underlying decision-making algorithms.

To address these concerns, recent research has begun to focus on enhancing the interpretability of deep learning models. For instance, the authors of \cite{chrosniak2024deep} introduced a physics guard layer to constrain the learned parameters within physical bounds. Similarly, \cite{Zhou2024} combined physical prior knowledge with data-driven modules to learn the vehicle lateral dynamics, while \cite{Wegrzynowski2024} adopted neural tire models. 

Following a similar research line, the authors of \cite{da2020modelling,da2020mental,piccinini2023physics,Piccinini2024_ard,Piccinini_ABS_2025,DaLio_SideSlip_2023,Pagot2023} introduced Model-Structured Neural Networks (MS-NNs) to embed the system's dynamics laws inside their architectures, for modeling \cite{da2020modelling,da2020mental,Piccinini_ABS_2025}, control \cite{da2020mental,piccinini2023physics,Piccinini2024_ard,Pagot2023} and estimation \cite{DaLio_SideSlip_2023} of the vehicle dynamics. Compared to general-purpose NNs, these physics-driven models showed enhanced interpretability and generalization from limited training data.

To the best of our knowledge, the existing examples of NN-based steering controllers for autonomous vehicles are limited by at least one of the following:
\begin{itemize}
    \item MS-NNs were used as steering controllers in simulation \cite{piccinini2023physics,Piccinini2024_ard}, but not with real-world data.
    \item The neural steering controllers had general-purpose architectures and required large training sets \cite{piccinini2023predictive,kuutti2020survey}.
    \item The neural steering controllers were conceived for urban driving and not for high-performance racing \cite{da2020mental}.
    % \item The lack of validation using real-world measurements from on-track vehicle testing;
    \item Absence of open-source frameworks to easily replicate the proposed implementations.
\end{itemize}

\subsection{Contributions}
To address the previous limitations, this paper's contributions are the following:
\begin{itemize}
    \item We present a new model-structured neural network for vehicle steering control, named MS-NN-steer, by extending the work of \cite{piccinini2023physics}. MS-NN-steer explicitly models the influence of the vehicle velocity on the steering characteristics: we show that this is a non-negligible effect in high-performance racing.
    \item We train and validate our MS-NN-steer using real-world data from the 2024 Abu Dhabi Autonomous Racing League (A2RL) competition. 
    \item The proposed MS-NN-steer is compared against three benchmarks: the steering controller used by the winning team of the A2RL competition, a general-purpose neural network (G-NN), and the baseline MS-NN of \cite{piccinini2023physics}.
    % conduct an extensive performance evaluation using real measurement data from racing vehicles. The proposed approach is benchmarked against two alternative models, demonstrating its superior accuracy and generalization capabilities;
    \item We release open-source our code to generate customized MS-NN architectures and to perform hyperparameters optimization. 
\end{itemize}

% Keeping in mind the presented limitation, our approach aims to identify the vehicle dynamic behavior by leveraging the underlying physical laws. To this end, we design a neural network architecture that mirrors the structure of the physical model, enabling the network to achieve high accuracy while preserving the physical interpretability of its parameters. This class of networks is referred to as Model-Structured Neural Networks (MS-NNs). The proposed architecture is validated using real-world data collected during the 2024 A2RL event held in Abu Dhabi.

Finally, the paper is organized as follows. The design of the proposed feedforward steering control is detailed in Section II, while the benchmarks are in Section III.
The analysis of the results is performed in Section IV, and Section V draws the conclusions.

\section{Steering Control Design}\label{control_architecture}
\subsection{Overview}
\begin{figure}[!t]
\vspace{5pt}
    \centering
    \includegraphics[width=\columnwidth]{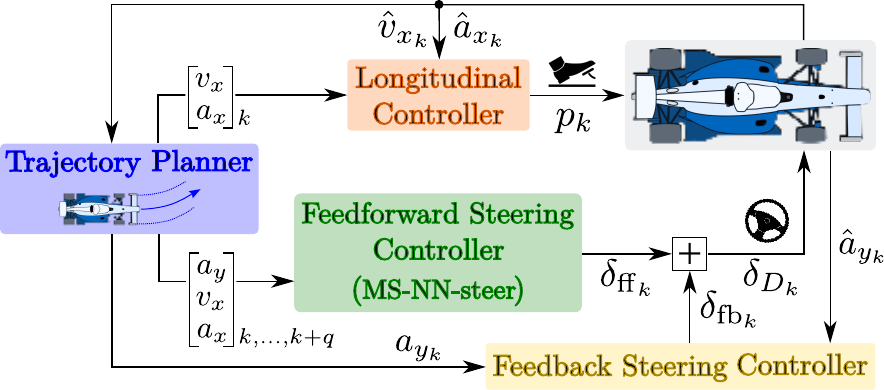}
    \caption{Trajectory planning and control framework on the A2RL race car. The MS-NN-steer of this paper is a feedforward steering controller (green block), computing the steering angle to execute the planned trajectories.}
    \label{fig:overview}
\end{figure}

Fig. \ref{fig:overview} overviews the trajectory planning and control framework of the TUM Autonomous Motorsport software stack \cite{betz2023tum}, which was employed in the A2RL competition. In this architecture, the trajectory planner uses a Tube Model Predictive Controller (Tube-MPC) to optimize the vehicle maneuver on a receding horizon. The planned trajectory is then executed by feedforward-feedback steering controllers and a longitudinal controller. 

% In modular autonomous driving architectures, trajectory tracking is commonly employed to let the vehicle execute the trajectory planned, either online or offline, by a higher-level motion planner. In our approach, the higher layer of the trajectory tracking module leverages a Tube Model Predictive Controller (Tube-MPC), based on a limited-friction point-mass vehicle model, to compute reference signals for the lower control layers, specifically the longitudinal and steering controllers.\\
This paper develops a new neural network-based feedforward steering controller. The main methodological contribution is the design of a Model-Structured Neural Network (MS-NN), which integrates prior knowledge of the nonlinear vehicle dynamic laws into its architecture. 

We consider two incremental versions of our MS-NN. The first version, named MS-NN-base and derived from \cite{piccinini2023physics}, is our baseline, which models the combined lateral vehicle dynamics considering the effect of the longitudinal acceleration $a_x$. The second variant, MS-NN-steer, is a new extension of MS-NN-base and captures the influence of the longitudinal speed $v_x$ on the vehicle's steering characteristics. 

\subsection{MS-NN: Inputs and Output} \label{sec:IO}

%Using the terminology of \cite{da2020mental,piccinini2023physics}, our MS-NN is an inverse dynamic model, computing the steering angle $\delta_k$ to track the trajectories planned by the Tube-MPC. 
As shown in Fig. \ref{fig:overview}, our MS-NN computes the steering angle $ \delta_{k}$, where $k \in \mathbb{N}$ is the index of the current time step. The MS-NN's inputs are the trajectories planned by the Tube-MPC. Specifically, the inputs are windows of future planned values for the lateral and longitudinal accelerations $\{a_y,a_x\}$, and the longitudinal speed $v_x$:
\begin{subnumcases}{\label{eq_vectors_future_inputs}}
	\boldsymbol{a_y}_k = \begin{bmatrix} a_{y_k}, \dots, a_{y_{k+q}} \end{bmatrix} \label{eq_future_ay}	 \\
	\boldsymbol{a_x}_k = \begin{bmatrix} a_{x_k}, \dots, a_{x_{k+q}} \end{bmatrix} \label{eq_future_ax} \\
    \boldsymbol{v_x}_k = \begin{bmatrix} v_{x_k}, \dots, v_{x_{k+q}} \end{bmatrix}  \label{eq_future_vx}
\end{subnumcases}
where $q$ is the number of future time steps, the time step size is $T$, and both $q$ and $T$ are tunable parameters.

% These inputs include the predicted lateral and longitudinal accelerations (${a_y, a_x}$) as well as the longitudinal velocity ($v_x$), so  $ \delta_{k} = F_N(a_{y,k},a_{x,k},v_{x,k})$.
% The vectors $a_{y,k}, a_{x,k},v_{x,k}$ represent the present and future predictions, with the subscript $k \in \mathbb{N}$ denoting the current time step, and $x_k = x(kT_{sNN})$ indicating the quantity $x$ at time $t = kT_{sNN}$.\\

\subsection{MS-NN-base: Baseline Version}\label{MS-NN-base}
The baseline version of our MS-NN, named MS-NN-base, is derived from \cite{piccinini2023physics}, as is here briefly recalled. The internal architecture of MS-NN-base is shown in Fig. \ref{fig:extended}, \textit{without} the red term $k_y(v_{x})$ inside the function $G(\cdot)$. This model is expressed as:
\begin{equation}
    \delta_{k} = \sum_{j=1}^{n_v}\sum_{l=1}^{n_x} \Big( \mathbf{G}( \boldsymbol{a_y}_k, \boldsymbol{v_x}_k,\boldsymbol{a_x}_k) \odot \, \boldsymbol{\phi}_{jl}(\boldsymbol{v_x}_k,\boldsymbol{a_x}_k) \Big) \cdot 
    \begin{bmatrix}
    F_{jl_1} \\
    \vdots \\
    F_{jl_{q+1}}
    \end{bmatrix}
\label{eq:MS-NN-base}
\end{equation}
where $\odot$ denotes the Hadamard element-wise 
product. The structure of MS-NN-base consists of two main parts. The first part learns the \textit{quasi steady-state} lateral vehicle dynamics, while the second part captures the transient dynamics. 

\subsubsection{Quasi Steady-State Vehicle Behavior}

The first part of MS-NN-base is the vector function $\mathbf{G}(\cdot)$ (blue block in Fig. \ref{fig:extended}), which is designed to learn the \textit{quasi steady-state} lateral dynamics of the vehicle. The term \textit{quasi} refers to the fact that the vehicle's dynamics are not strictly steady-state, as the longitudinal acceleration $a_x$ can be non-zero. As shown in its expanded view in Fig. \ref{fig:extended}, 
$\mathbf{G}(\cdot)$ resembles a neuro-fuzzy system \cite{nelles2000local,nelles2020nonlinear}, and is composed of $n_y\cdot n_x$ local models, each of which learns the dynamics in a predefined range of $a_y$ and $a_x$.  

\begin{figure}[!t]
    \centering
    \includegraphics[width=\columnwidth]{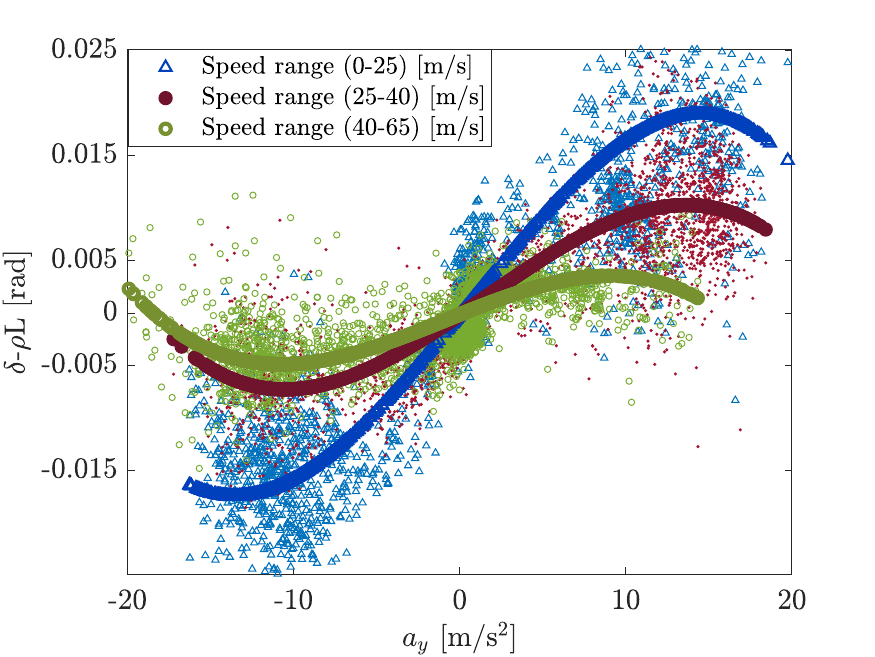}
    \caption{Handling diagram (HD) derived from the real telemetry data of the A2RL car. The plot highlights the dependency of the HD on the vehicle speed $v_x$, which is a crucial physical phenomenon that we aim to capture in our extended MS-NN-steer architecture.
    %: Vehicle operational point at different speed ranges during the Abu-Dhabi track run. 
    The solid lines are polynomial fits of the data points.
    %A third-degree polynomial fit was applied to model the nonlinear trend in the measurements.
    }
    \label{fig:HD}
\end{figure}

To describe the internal design of $\mathbf{G}(\cdot)$, we first recall the handling diagram (HD), depicted in Fig. \ref{fig:HD}. The HD is a representation of a vehicle's steady-state nonlinear understeering and oversteering characteristics. It is generated by plotting the solution to the following equation:
\begin{equation}
    \delta - \rho L = \delta - \frac{a_y}{v_x^2} L = \alpha_1 - \alpha_2
    \label{eq:example}
\end{equation}
where $L$ is the vehicle wheelbase, and $\{\alpha_1,\alpha_2\}$ are the side slip angles of the front and rear tires.
The HD quantifies the mismatch between the real steering angle $\delta$ (the average angle at the front wheels) and the kinematic steering angle $\rho L$, with $\rho = a_y/v^2_x$ being the trajectory curvature. 

As depicted in Fig. \ref{fig:extended}, the function $\mathbf{G}(\cdot)$ in \eqref{eq:MS-NN-base} consists of local neuro-fuzzy models $g_{il}(\cdot)$, where $i \in \{1,\dots,n_y\}$ and $l \in \{1,\dots,n_x\}$ are the indices of the local models. To design $g_{il}(\cdot)$ using \eqref{eq:example}, we need a relation between $\{ \alpha_1,\alpha_2\}$ and the accelerations $\{a_y,a_x\}$ in the vicinity of a local model's center $\{ a_y=a_{y_{0_i}}, a_x=a_{x_{0_l}}\}$. As shown in \cite{piccinini2023physics}, this relation can be obtained by locally approximating the equations of a nonlinear double-track vehicle model, around $\{ a_y=a_{y_{0_i}}, a_x=a_{x_{0_l}}\}$. 
After performing the mathematical derivation outlined in \cite{piccinini2023physics}, the steering angle $\delta$ is locally expressed as follows:
\begin{multline}\label{eq:Base_g}
    g_{il}(a_y,v_x,a_x) \\
    = \frac{a_y}{v_x^2} L + k_{y1_i} \mathrm{sign}(a_y) + k_{y2_i} (a_y-a_{y,0_i}\mathrm{sign}(a_y)) + \\ 
    k_{y3_i}k_{x1_l}\cdot(a_y-(a_{y0_i}+k_{y4_i})\mathrm{sign}(a_y))(k_{x2_l}+a_x+a_{x0_l})\cdot\\
    [1+k_{y5_i}(a_y-a_{y0_i}\mathrm{sign}(a_y))\\
    +k_{x3_l}\cdot(a_x-a_{x0_l})+k_{x4_l}(a_x-a_{x0_l})^2+\\
    k_{y6_i}k_{x5_l}(a_y-a_{y0_i}\mathrm{sign}(a_y))(a_x-a_{x0_l})]
\end{multline}
The function $ \mathbf{g}_{il}(\boldsymbol{a_y}_k, \boldsymbol{v_x}_k,\boldsymbol{a_x}_k)$ in Fig. \ref{fig:extended} is the vector form of \eqref{eq:Base_g}, which is omitted for brevity. Each of the local models $g_{il}(\cdot)$ has $6 + 5 = 11$ learnable parameters, namely the set $\{k_{y1_i} , \dots, k_{y6_i}, k_{x1_l} ,\dots, k_{x5_l} \}$.
%
% The steady-state lateral dynamics is mainly captured by the vector function $\mathbf{G}(a_{y_k}, v_{x_k},a_{x_k})$, which learns $n_y\cdot n_x$ local models of the vehicle's handling characteristics, where $n_y$ and $n_x$ are the chosen number of local model in the range of $a_y$ and $a_x$ respectively. 
%
The local models $\mathbf{g}_{il}(\cdot)$ are activated by the vector functions $\boldsymbol{\phi}_{il}(\boldsymbol{a_y}_k,\boldsymbol{a_x}_k) = \boldsymbol{\phi}_{i}(\boldsymbol{a_y}_k)\odot \boldsymbol{\phi}_{l}(\boldsymbol{a_x}_k)$ with $i \in \{1,\dots,n_y\}$ and $l \in \{1,\dots,n_x\}$ (see Fig. \ref{fig:extended}).

$\mathbf{G}( \boldsymbol{a_y}_k, \boldsymbol{v_x}_k,\boldsymbol{a_x}_k)$ returns a vector of $q + 1$ future steady-state steering angle predictions.

\subsubsection{Transient Vehicle Dynamics}

From the classical theory of vehicle dynamics \cite{Guiggiani2018}, the transient lateral dynamics depend on the vehicle speed $v_x$ and acceleration $a_x$. In MS-NN-steer, these dynamics are captured by the fully connected layers $\boldsymbol{F}_{jl}$ (green block in Fig. \ref{fig:extended}), which learn linear combinations of the $\mathbf{G}(\cdot)$ vector entries across different ranges of $v_x$ and $a_x$. This is done using the activation functions $ \boldsymbol{\phi}_{jl}(\boldsymbol{v_x}_k,\boldsymbol{a_x}_k)$, where $l \in \{1,\dots,n_x\}$, $j \in \{1, \dots, n_v\}$, and $n_v$ is the number of local models in the range of $v_x$.

%From the classical theories of vehicle dynamics, we know that the transient lateral dynamics change as a function of the vehicle speed $v_x$.
%In MS-NN-steer, the transient lateral dynamics are captured by the green block in Fig. \ref{fig:extended} through the fully connected layers $F_{jl}$, which learn the transient dynamics in different ranges of $v_x$, where $j \in \{1, \dots, n_v\}$, $l \in \{1,\dots,n_x\}$, and $n_v$ is the number of local model in the range of $v_x$. These layers learn linear combinations of the entries of the $\mathbf{G}(\cdot)$ vector across different ranges of $v_x$ and $a_x$, using the activation functions $ \boldsymbol{\phi}_{jl}(\boldsymbol{v_x}_k,\boldsymbol{a_x}_k)$.

\subsection{MS-NN-steer: Extended Version}

\begin{figure}[]
\vspace{5pt}
    \centering
    \includegraphics[width=\columnwidth]{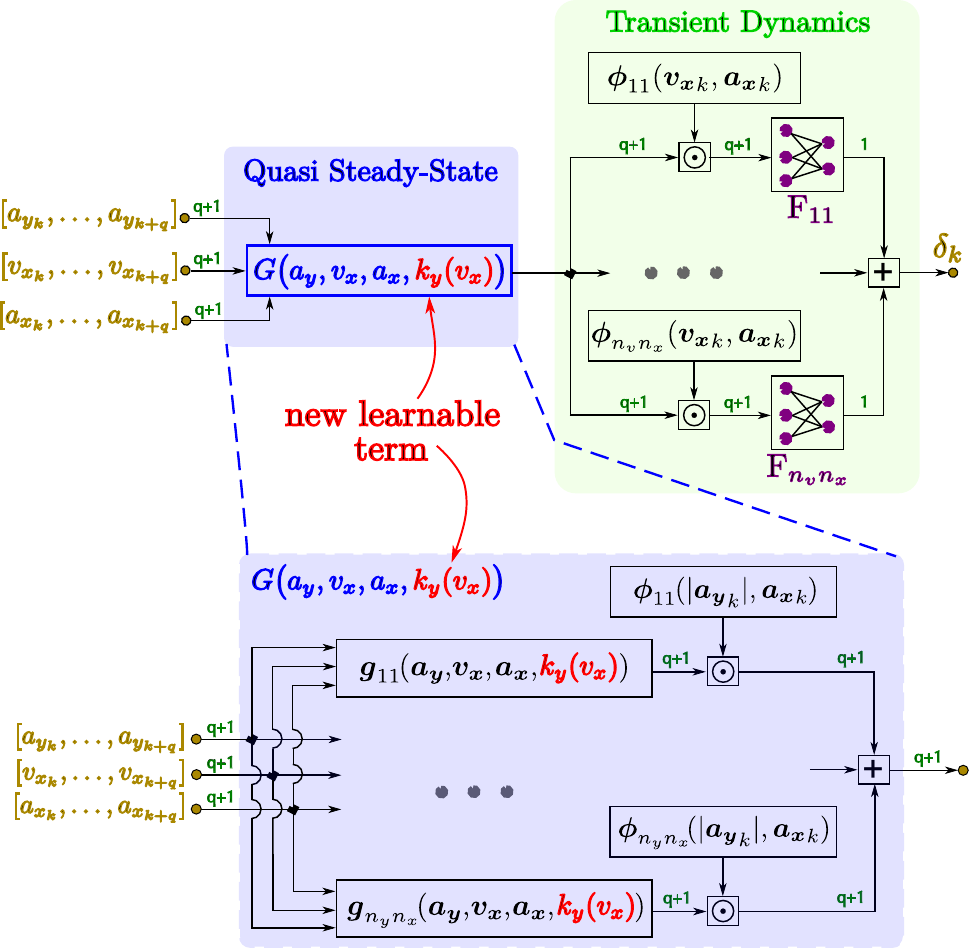}
    \caption{Proposed MS-NN-steer architecture. The red learnable term $k_y(v_{x})$ is part of the novelty of this work, as it captures the influence of the vehicle speed $v_x$ on the steering characteristics and the handling diagram (Fig. \ref{fig:HD}).}
    \label{fig:extended}
\end{figure}

The methodological novelty of this work lies in the improvement of the $g_{il}(\cdot)$ function in Fig. \ref{fig:extended} and equation \eqref{eq:Base_g}. To illustrate this, let us examine the experimental handling diagram shown in Fig. \ref{fig:HD}, which displays three different third-degree polynomial fitted curves corresponding to distinct speed ranges. The shape of the handling diagram varies as a function of the vehicle speed, both in the linear region (i.e., for low $a_y$) and in the nonlinear region. 
We argue that this peculiar shape of the diagram is due to the speed-dependent aerodynamics and handling characteristics of our racing car, which was not accounted for in the MS-NN-base of Section \ref{MS-NN-base}. 

To address this, we introduce an extended version of the local models $g_{il}(\cdot)$ in \eqref{eq:Base_g}, where some of the parameters $k_y$ become now functions of the speed. Specifically, to capture the effect of $v_x$, we extend the local models by replacing the parameters $\{k_{y1_i}, k_{y2_i}\}$ with learnable functions $\{k_{y1_i}(v_x),  k_{y2_i}(v_x)\}$:
\begin{multline}
    g_{il}\big(a_y,v_x,a_x,\overbrace{k_{y}(v_x)}^{\text{new term}}\big) \\
    = \frac{a_y}{v_x^2} L + \overbrace{k_{y1_i}(v_x)}^{\text{new term}} \mathrm{sign}(a_y) + \overbrace{k_{y2_i}(v_x)}^{\text{new term}} \, (a_y-a_{y,0_i}\mathrm{sign}(a_y)) + \\ 
    k_{y3_i}k_{x1_l}\cdot(a_y-(a_{y0_i}+k_{y4_i})\mathrm{sign}(a_y))(k_{x2_l}+a_x+a_{x0_l})\cdot\\
    [1+k_{y5_i}(a_y-a_{y0_i}\mathrm{sign}(a_y))\\
    +k_{x3_l}\cdot(a_x-a_{x0_l})+k_{x4_l}(a_x-a_{x0_l})^2+\\
    k_{y6_i}k_{x5_l}(a_y-a_{y0_i}\mathrm{sign}(a_y))(a_x-a_{x0_l})]
    \label{eq:Extended_g}
\end{multline}
The functions $k_{y1_i}(v_x)$ and $k_{y2_i}(v_x)$ in \eqref{eq:Extended_g} are modeled as learnable polynomials of the form: 
\begin{align*}
    k_{y1i}(v_x)=\sum_{s=0}^{n_{p_1}}c_{1_{is}}v^s_x &\qquad  k_{y2_i}(v_x)=\sum_{w=0}^{n_{p_2}}c_{2_{iw}}v^w_x
\end{align*}
where ${c_{1_{is}}, c_{2_{iw}}}$ are learnable parameters. The polynomial degrees $\{n_{p_1}, n_{p_2}\}$ for $k_{1_i}(v_x)$ and $ k_{2_i}(v_x)$ are optimized through experimental validation, which results in $n_{p_1}=3$ and $n_{p_2}=1$.

\begin{comment}
\begin{equation}
\begin{aligned}
&\left\{
\begin{aligned}
&\boldsymbol{\bar{g}}_i(\boldsymbol{a}_{y,k}, \boldsymbol{v}_{x,k}) = \boldsymbol{a}_{y,k} \odot (\boldsymbol{v}_{x,k})^{\circ -2} L + k_{1_i} \operatorname{sign}(\boldsymbol{a}_{y,k}) \\
& \quad + k_{2_i} (\boldsymbol{a}_{y,k} - a_{y0_i} \operatorname{sign}(a_{y,k})) \\
&\boldsymbol{\bar{G}}(\boldsymbol{a}_{y,k}, \boldsymbol{v}_{x,k}) = \sum_{i=1}^{n_y} \left( \boldsymbol{\bar{g}}_i(\boldsymbol{a}_{y,k}, \boldsymbol{v}_{x,k}) \odot \boldsymbol{\bar{\phi}}_i (|\boldsymbol{a}_{y,k}|) \right)
\end{aligned}
\right.
\end{aligned}
\end{equation}
\end{comment}

% \begin{figure}[!t]
%     \centering
%     \subfloat[]{\includegraphics[width=\columnwidth]{Figure/NN_local_linear_models_Ax_N_channels_new_2.pdf}\label{fig:extended}} \\
%     \subfloat[]{\includegraphics[width=\columnwidth]{Figure/NN_thread_layers_with_Ax.pdf}\label{fig:extended2}}\\
%     \subfloat[]{\includegraphics[width=\columnwidth]{Figure/GP_ff_NN_benchmark.pdf}\label{fig:general}}\\
%     \caption{%\revM{Evidenzia transient e steady-state, espandi G, e scrivi che $k_y$ è la novità.} 
%     Neural Networks: $(a)$ ; $(b)$ ; $(c)$.}
% \end{figure}

\begin{figure}[]
    \centering
    \includegraphics[width=\columnwidth]{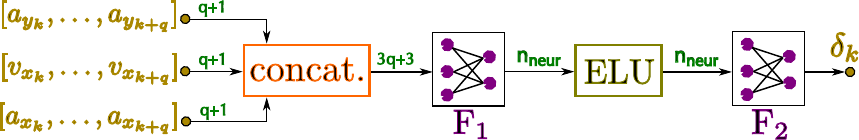}
    \caption{General-purpose neural networl (G-NN) used as a benchmark.}
    \label{fig:general}
\end{figure}

\section{Benchmarks}\label{benchmarks}

Besides the MS-NN-base of \cite{piccinini2023physics}, we benchmark our new MS-NN-steer against the following techniques.

\subsection{General-Purpose Neural Network}\label{GN}

The first benchmark is a general-purpose neural network (G-NN), shown in Fig. \ref{fig:general}. G-NN has the same inputs and output as our MS-NN-steer. The input vectors are first concatenated and then passed through a two-layer feedforward NN, with an ELU (exponential linear unit) activation function. The first layer has $n_{\mathrm{neur}}$ neurons, while the second layer has $1$ neuron, which produces the scalar output $\delta_k$. 
% fully connected layer. This layer produces $n_{neur}$ outputs , which are activated using the ELU function. These outputs are subsequently processed by a second fully connected layer, which yields the final steering angle $\delta_k$. 
The total number of learnable parameters in G-NN nearly matches the one of our MS-NN-steer. This allows for a fair comparison, and will highlight the performance and generalization advantages of MS-NNs.

\subsection{Baseline A2RL Steering Controller}\label{current_controller}
As an additional benchmark, we use the feedforward steering controller of the TUM Autonomous Motorsport software stack \cite{betz2023tum} that won the A2RL competition in 2024 (Fig. \ref{fig:A2RL}). This controller, hereafter named A2RL-control, computes the steering angle $\delta_k$ as a sum of four terms:
\begin{equation}
    \delta_{k}=\delta_{\mathrm{kin}}(a_{y_k},v_{x_k})+\delta_{\mathrm{us}}(a_{y_k})+\delta_{\mathrm{ax}}(a_{y_k},a_{x_k})+\delta_\mathrm{off}
    \label{eq:A2RL_control}
\end{equation}
where $\delta_{\mathrm{kin}} = a_{y} \, L/v_{x}^2$ is the kinematic steering angle. $\delta_{\mathrm{us}}$ considers the understeering behavior and uses a first-order dynamics to produce an additional steering angle $\delta_{\mathrm{us}_k} = \delta_{\mathrm{us}_{k-1}}+(K_\mathrm{us} \, a_{y_k} - \delta_{\mathrm{us}_{k-1}}) \, \frac{\Delta t}{T_\mathrm{us}}$, where $K_\mathrm{us}$ is a tunable understeering gradient, $\Delta t$ is the sampling time, and $T_\mathrm{us}$ is the tunable time constant of the first-order dynamics. $\delta_{\mathrm{ax}}$ is a compensation term that accounts for the weight transfers and aerodynamic effects, and is modeled as $\delta_{\mathrm{ax}_k}= \big(\delta_{\mathrm{ax}_{k-1}} + (K_\mathrm{ax} \, a_{x_k} - \delta_{\mathrm{ax}_{k-1}}) \, \frac{\Delta t}{T_\mathrm{ax}}\big) \, a_{y_k}$, where $K_\mathrm{ax}$ and $T_\mathrm{ax}$ are tunable parameters. 
Finally, $\delta_\mathrm{off}$ is a static offset due to asymmetries in the car setup.

% To improve tracking performance in presence of understeer the algorithm applies a delayed supplementary steering angle dependent on the requested lateral acceleration with a first-order low pass. The longitudinal acceleration dependence is modeled by first-low pass on longitudinal acceleration that is then multiplied with the requested lateral acceleration. 

\section{Results with Experimental Data}
\label{exp}

% \subsection{Overview}

% An in-depth analysis was carried out using the experimental data presented in the following subsection to assess the performance of the proposed architecture. \\
% Then the discussion is organized into fourth subsections: C. presents the nndoely framework exploited to generate all neural networks discussed in this paper as well as the training and overall analysis; D. shows the analysis that guided the selection of the hyperparameters for our proposed neural network; E. investigates the performance comparison between our network and the base version \ref{MS-NN-base} as well as the two benchmarks introduced in Section. \ref{benchmarks}; F. analyzes the evaluation of networks performance under several different training hyperparameter conditions.\\

\subsection{Real-World Autonomous Race Car Dataset}\label{dataset}

\subsubsection{A2RL Autonomous Race Car}
The test platform, EAV24, is an autonomous open-wheel race car based on a Dallara Super Formula SF23 chassis. The telemetry data used in this paper was collected during the 2024 Yas Marina Event of the A2RL racing series. This was an autonomous race between eight teams running their cars in single-vehicle fastest laps and multi-vehicle racing sessions, with speeds up to 260 km/h. The training datasets of this paper contain the
%For preliminary open-loop training and evaluation of the proposed feedforward control concepts, we use 
velocity and acceleration signals computed by a state estimator, which fuses IMU, GNSS, LiDar, and Radar sensor signals \cite{betz2023tum}. The steering angle signal is measured by the steer-by-wire system of the car.
% \textcolor{Green}{TO DO (Georg): describe the car (top speed etc), the competition (very briefly) and the sensor setup. In particular, say how the signals needed by the MS-NN (vehicle speed, long/lat accelerations, steering angle) are measured.}

\subsubsection{Training, Validation, and Test Datasets} 

\begin{figure}[!t]
\vspace{5pt}
    \centering
    \includegraphics[width=\columnwidth]{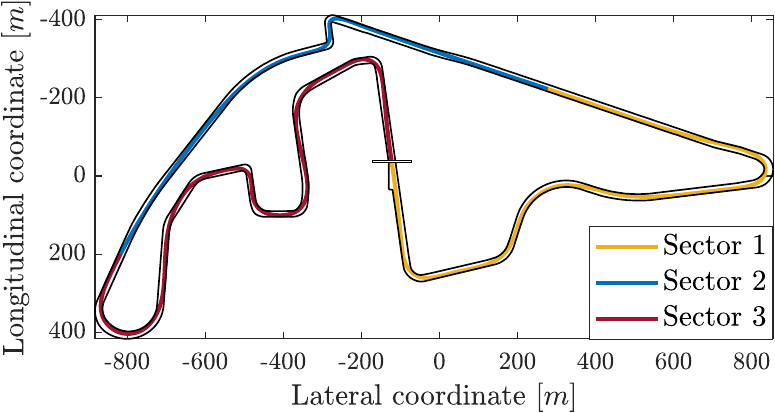}
    \caption{Yas Marina track sectors used to generate training and validation datasets with different sizes, for the neural network controllers of this paper.}
    \label{fig:track_segments}
\end{figure}

Fig. \ref{fig:track_segments} shows the racetrack layout with the sectors considered for the generation of our datasets. The data in this paper originated from two pre-qualifying laps. 
% Consequently, the Training and Validation datasets were constructed using data from the first two laps. 
% The proposed solution, as well as the benchmarks, were then tested on faster laps, specifically evaluating the performance on the third, fourth, and fifth laps, while the sixth lap was excluded as it corresponded to the cool-down phase and the return to the pits. 

As reported in Table \ref{tab:datasets}, the training set uses the first lap, while the validation set uses the second one. The reference speed was progressively increased from the first to the second lap, and the vehicle had different tire sets in each lap. Thus, the validation set is used to assess the generalization potential of the neural controllers in unseen conditions. 
To further evaluate the capability to learn with limited data, we generate three different training sets of increasing size, as reported in Table \ref{tab:datasets}. 

\begin{table}[]
    \center
    \caption{Training and validation datasets, using the track sectors in Fig. \ref{fig:track_segments}.}
    \begin{tabular}{ccccc}
        \toprule
        & \multicolumn{3}{c}{\textbf{Training sets}} & \textbf{Validation set} \\
        \cmidrule(lr){2-4} \cmidrule(lr){5-5} & Small & Medium & Large & \\ 
        \midrule
        \begin{tabular}{@{}c@{}}Track\\sectors\end{tabular} & 3 & 1+3 & 1+2+3 & 1+2+3 \\
        \midrule
        Lap & 1 & 1 & 1 & 2 \\
        \toprule
    \end{tabular}
    \label{tab:datasets}
\end{table}

    % 
    %\cmidrule(lr){2-4} \cmidrule(lr){5} \cmidrule(lr){6} 
    %\multicolumn{1}{c}{} & Small & Medium & Large &  & \\
    %\midrule
      %  \begin{tabular}{@{}c@{}}Track\\sectors\end{tabular} & 3 & 1+3 & 1+2+3 & 3+1 & 1+2+3 \\
      %  \toprule
  %  \caption{Training and validation sets, using the track sectors in Fig.\ref{fig:track_segments}.}
 %   \label{tab:datasets}
%\end{table}

\begin{figure*}[!t]
\vspace{5pt}
    \centering
    \includegraphics[width=\textwidth]{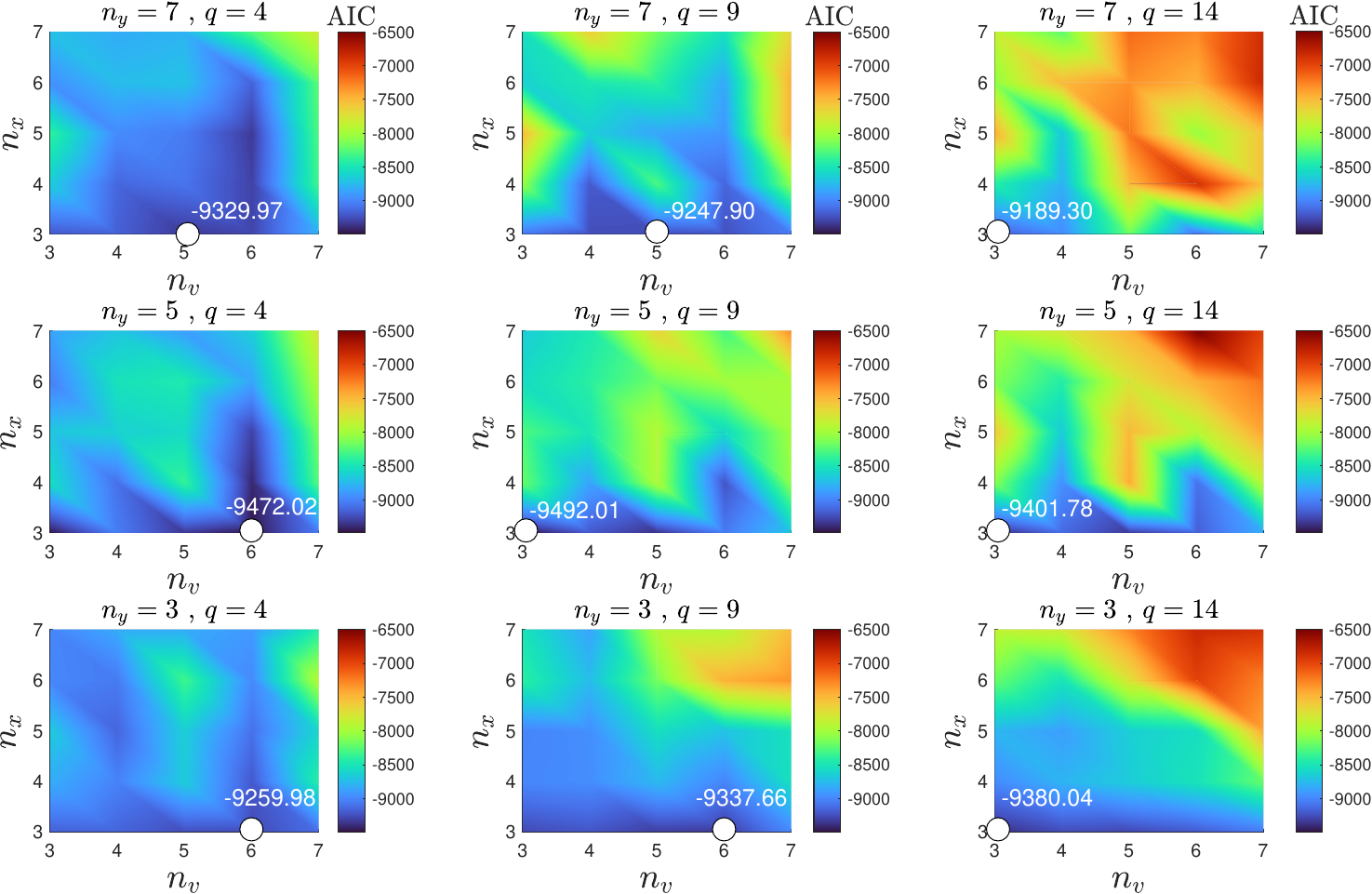}
    \caption{Grid search to tune the MS-NN-steer's hyperparameters. The figure shows the Akaike Information Criterion (AIC) for each combination of $(q,n_y,n_x,n_v)$ in their ranges. The AIC metric is a trade-off between model complexity, quantified by the number of learnable parameters, and the accuracy on the validation dataset. The lowest value, i.e., the most sample-efficient model, results in $q=9$, $n_y=5$, $n_x=3$ and $n_v=3$.}
    % Hyperparameters Network Analysis: AIC KPI for each combination of $(q,n_y,n_x,n_v)$ in the ranges assumed. The AIC metric indicates the trade-off between model complexity, quantified by the number of learnable parameters, and the accuracy achieved on the validation dataset.
    % The lowest value, i.e. the best balance between model simplicity and predictive performance, results in $q=9 \, , n_y=5 \, , n_x=3$ and $n_v=3$. 
    \label{fig:hyperparameters}
\end{figure*}

\subsection{MS-NN Implementation and Training}\label{nnodely}

Designing highly structured neural networks (e.g., MS-NNs) poses challenges when using general-purpose machine learning frameworks like PyTorch or TensorFlow. Although powerful and flexible, these libraries typically demand substantial expertise to embed domain-specific structures and physical priors, leading to a complex and time-consuming development process.\\
To address this challenge, we used \texttt{nnodely}, a new open-source software tool that streamlines the implementation of MS-NNs, available at \url{https://github.com/tonegas/nnodely.git}. \\
\texttt{nnodely} provides an intuitive and modular environment tailored for rapid prototyping and evaluation of MS-NN architectures. It lowers the entry barrier for users with classical modeling backgrounds (e.g., engineers, physicists) by streamlining the integration of physical principles and facilitating deployment.

The training of the proposed and benchmark neural models is conducted in \texttt{nnodely}, using supervised learning to minimize the Root Mean Square Error (RMSE) between the predicted steering angles $\delta$ and the corresponding measured values $\delta_{\mathrm{meas}}$, obtained from the experimental data. Training is performed using the Adam algorithm. To prevent overfitting, we employ the early stopping technique: the training process is stopped if the validation loss does not improve after a predefined number of epochs.\\
The training hyperparameters are: learning rate $10^{-3}$, $8000$ epochs, batch size $1000$, and early stopping with a patience of $1500$ epochs.

% In particular in this work we exploit the local model features for creating the function $G$ (and not only) \commM{Please check} and the sample-time windows of the signals for managing the network memory avoiding a state space approach (this solution is demonstrated to be the best for a finite memory system \cite{da2020modelling}). 

\subsection{Evaluation Metrics} \label{sec:metrics}

In this paper, we consider the following key performance indicators.
The Root Mean Square Error (RMSE) is used to quantify the model accuracy, while the Fraction of Variance Unexplained (FVU) quantifies the proportion of variance in the target data that remains unexplained by the model (lower FVU values indicate better performance). Finally, the Akaike Information Criterion (AIC) \cite{bozdogan1987model} is a metric for model selection which trades off the model accuracy versus its complexity (number of learnable parameters), with lower AIC values indicating more sample-efficient models.

\subsection{Hyperparameters Tuning}\label{Hyp_net}

We tune the MS-NN-steer's hyperparameters with a grid-search approach, to minimize the AIC metric (Section \ref{sec:metrics}) on the validation dataset. We consider the following variables: the number of future time steps $q$ in the input vectors, the numbers of local models $\{n_y,n_x,n_v\}$ for the lateral accelerations, the longitudinal accelerations and the longitudinal speed, respectively. The following ranges are assumed: $q \in (4,9,14)$, $n_y \in (3,5,7)$, $n_x \in (3, \dots, 7)$ and $n_v \in (3, \dots, 7)$. 

The results are shown in Fig. \ref{fig:hyperparameters}, which displays a heatmap of the AIC values computed for each combination of hyperparameters. 
% The AIC metric serves as an indicator of the trade-off between model complexity, quantified by the number of learnable parameters, and the accuracy achieved on the validation dataset. Lower AIC values suggest a more favorable balance between model simplicity and predictive performance.\\
Each subplot includes a white dot that marks the local minimum of the AIC within the corresponding parameter configuration. By examining these local minima and identifying the lowest overall value across all configurations, the best model corresponds to the setting with $q=9$, $n_y=5$, $n_x=3$ and $n_v=3$. This configuration yields the lowest AIC value, and thus the best balance between model complexity and predictive performance. 

Note that the MS-NN-steer's input vectors span a time window of $q \cdot T$ seconds, where the sampling time $T$ is $0.05$ s in our implementation. Thus, in the best setting, the input vectors span $0.45$ s, which is enough to cover most of the time-scale of the lateral vehicle dynamics \cite{piccinini2023physics}.  

Finally, the number of learnable parameters for MS-NN-steer is:
\begin{multline}
    N_{\mathrm{pars}}=
    (n_{p_1}+1+n_{p_2}+1)\cdot n_y +\\ +k_{y3,\cdots,6}\cdot n_y + k_{x1,\cdots,5}\cdot n_x + (q+1)\cdot n_x \cdot n_v= \\=(4+1+1+1)\cdot5 +4\cdot5 + 5\cdot 3 + (9 + 1) \cdot 3 \cdot 3 = 160
\end{multline}
% Using the inverse formula for the same number of inputs gives the value of $n_{neur}$ for the general model which results in being equal to $5$.

\subsection{Performance Evaluation}\label{performance}
\begin{figure*}[!t]
\centering
    \subfloat[]{\includegraphics[width=0.48\textwidth]{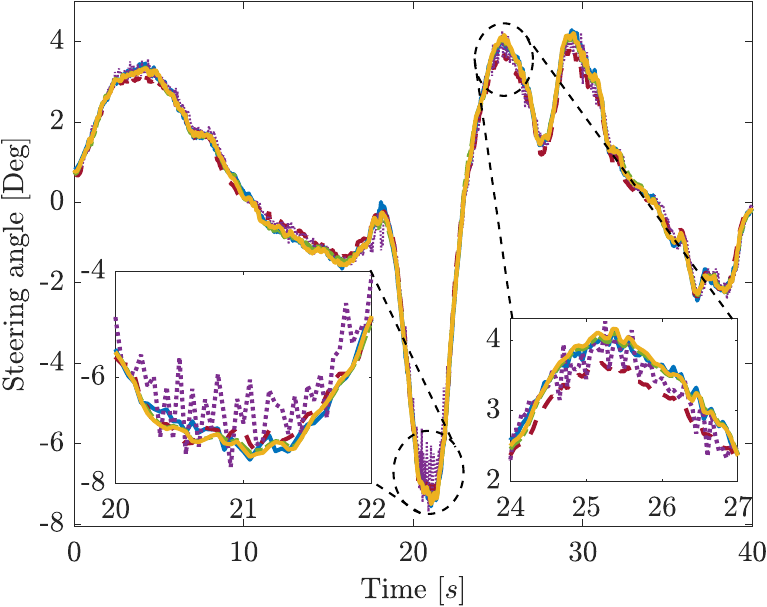}}\hfill%
    \subfloat[]{\includegraphics[width=0.48\textwidth]{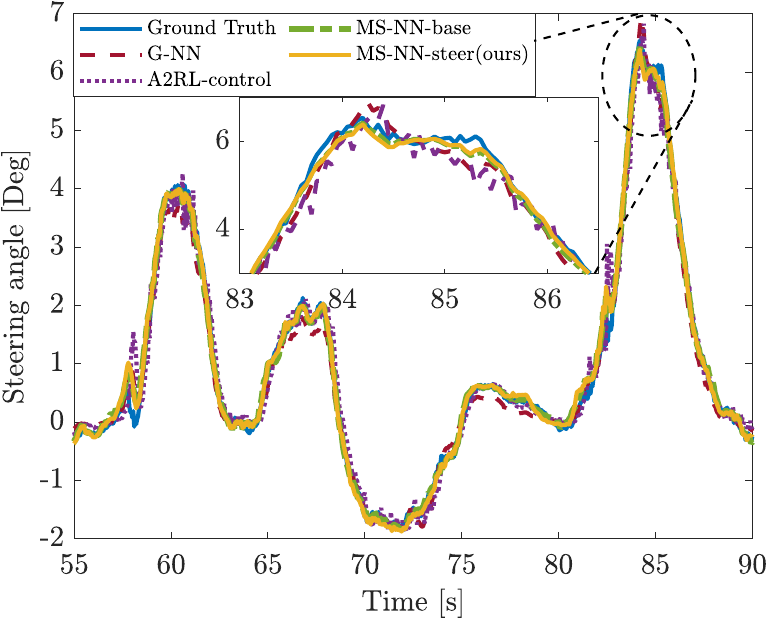}}\\
    %\subfloat[]{\includegraphics[width=0.49\textwidth]{Figure/Test_steer_2_210.pdf}}
    % \subfloat[]{\includegraphics[width=0.5\textwidth]{Figure/Box_errore.pdf}}
    \caption{Steering angle profiles computed by the proposed MS-NN-steer and the benchmarks, when trained with the large training set of Table \ref{tab:datasets}. Results on the training set (a) and validation set (b). Note that the plots show around $40$ s of the full datasets, for improved visibility.}
    \label{Steerresult}
\end{figure*}

\begin{figure}[]
    \centering
    \includegraphics[width=0.48\textwidth]{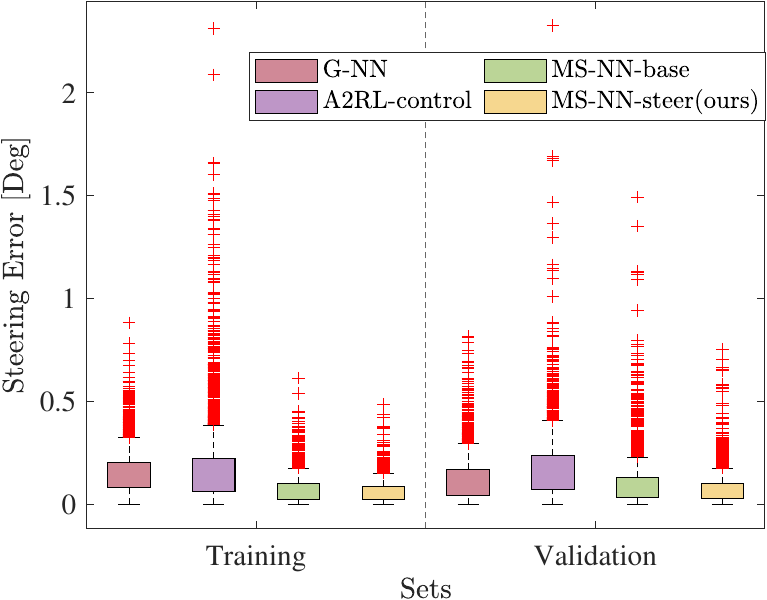}
    \caption{Boxplot of the steering angle errors, when using the large training dataset of Table \ref{tab:datasets}. The proposed MS-NN-steer outperforms the benchmarks both in training and validation, with lower mean, variance, and peak errors.}
    \label{fig:box_plot}
\end{figure}

\begin{table}[]
    \footnotesize
    \centering
    \caption{Results with the large training dataset.}
    \label{tab:rmse_fvu}
    \begin{tabular}{@{}cl*{4}{ccc}}
    \toprule %\midrule
     & Metrics & Training Set & Validation Set \\
    \midrule
    \multirow{2}{*}{G-NN} 
        & RMSE [deg] & $0.189$ & $0.217$ \\
        & FVU [-] & $0.0061$ & $0.0102$ \\
    \midrule
    A2RL-control
        & RMSE [deg] & $0.280$ & $0.259$ \\
        \cite{betz2023tum} & FVU [-] & $0.0130$ & $0.0141$\\
    \midrule
    MS-NN-base 
        & RMSE [deg] & $0.097$ & $0.109$ \\
        \cite{piccinini2023physics} & FVU [-] & $0.0023$ & $0.0022$ \\
    \midrule
    \textbf{MS-NN-steer} 
        & RMSE [deg] & $\textbf{0.079}$ & $\textbf{0.103}$ \\
        \textbf{(ours)} & FVU [-] & $\textbf{0.0012}$ & $\textbf{0.0016}$ \\
    %\midrule
    \bottomrule
    \end{tabular}
\end{table}
After tuning the numbers of parameters, the focus shifts to evaluating the model's accuracy and its ability to generalize on new data. Table \ref{tab:rmse_fvu}, Fig. \ref{Steerresult} and \ref{fig:box_plot} compare the performance of our MS-NN-steer and the benchmarks, using the large training set of Table \ref{tab:datasets}. The proposed MS-NN-steer outperforms all the benchmarks, improving the RMSE and FVU both on the training and validation sets. In validation, the performance gain over the G-NN, A2RL-control and MS-NN-base are $53\%$, $60\%$ and $6\%$ in terms of RMSE, respectively, while the FVU improves by $27\%$ compared to the MS-NN-base.
Also, MS-NN-steer reduces the peak steering errors by more than $60\%$ compared to the A2RL-control (Fig. \ref{fig:box_plot}). Finally, A2RL-control shows high-frequency steer oscillations in many parts of the dataset (Fig. \ref{Steerresult}(a)), while our MS-NN-steer yields smoother profiles and better robustness to noise on the input signals. Smoother steering profiles favor the vehicle's stability and reduce the risk of dangerous oversteering situations.
Overall, our MS-NN-steer can more accurately execute the planned trajectories, with a potentially large impact on the vehicle's overall performance and lap times.


% This is more evident when comparing our solution with the current A2RL control (Section~\ref{current_controller}). The proposed approach demonstrates a significant improvement in steering angle computation compared to the baseline controller. These improvements suggest a substantial gain in the vehicle’s stability and accuracy in trajectory following, which may have a notable impact on its overall race performance.\\
% To further validate the model’s effectiveness, Fig. \ref{Steerresult} (d) presents the boxplot of the error for both training and validation sets. The box height for the MS-NN-Extended model is consistently lower than that of the other models, indicating both a lower mean and standard deviation of the error.\\
% To quantify these improvements, Table \ref{tab:rmse_fvu} presents the values of RMSE and FVU, with a decrease of at least $10\%$ of RMSE and $50\%$ of FVU. \\

When trained with small datasets (Table \ref{tab:rmse_fvu_short}), the benchmark G-NN significantly degrades its performance on the validation set, with an RMSE of $0.327$ deg. In contrast, our MS-NN-steer achieves an RMSE of $0.097$ deg, thus demonstrating better generalization capabilities even with limited training data. 
This is a crucial aspect for real-world applications, where collecting data can be expensive and time-consuming, and the environment may change over time. 

Thus, the physics-based internal structure of our MS-NN-steer allows it to learn the underlying dynamics of the vehicle more effectively, reducing the risk of overfitting the training data and improving the performance on unseen scenarios. 

% This analysis is further extended by considering three different training dataset lengths, allowing for a more comprehensive evaluation of the model's performance across varying amounts of training data. Despite using shorter training datasets, the model continues to achieve excellent validation results, consistently outperforming the other networks.

\begin{table}[h!]
    \footnotesize
    \centering
    %\caption{Statistics results for medium and short Training and Validation datasets [Deg]: General Network (G-NN); MS-NN-Base; MS-NN-steer(ours).}
    \caption{Generalization performance using medium and small training datasets. The validation set is the same in all cases and the same as in Table \ref{tab:rmse_fvu}.}
    \label{tab:rmse_fvu_short}
    \begin{tabular}{@{}cl*{4}{ccc}}
    \toprule %\midrule
    \multicolumn{2}{c}{} & \multicolumn{2}{c}{Medium Train. Set} & \multicolumn{2}{c}{Small Train. Set} \\
    \cmidrule(lr){3-4} \cmidrule(lr){5-6} 
    \multicolumn{1}{c}{} & Metrics & Train. & Valid. & Train. & Valid. \\
    \midrule
    \multirow{2}{*}{G-NN} 
        & RMSE [deg] & $0.160$ & $0.229$ & $0.132$ & $0.327$ \\
        & FVU [-]  & $0.0059$ & $0.0150$ & $0.0025$ & $0.0792$ \\
    \midrule
    MS-NN-base 
        & RMSE [deg] & $0.079$ & $0.097$ & $0.069$ & $0.115$ \\
        \cite{piccinini2023physics} & FVU [-] & $0.0011$ & $0.0021$ & $0.0006$ & $0.0026$ \\
    \midrule
    %\multirow{2}{*}{\textbf{MS-NN-steer(ours)}} 
    \textbf{MS-NN-steer}
        & RMSE [deg] & $\textbf{0.068}$ & $\textbf{0.091}$ & $\textbf{0.063}$ & $\textbf{0.097}$ \\
    \textbf{(ours)} & FVU [-] & $\textbf{0.0008}$ & $\textbf{0.0018}$ & $\textbf{0.0006}$ & $\textbf{0.0020}$ \\
    %\midrule
    \bottomrule
    \end{tabular}
\end{table}

\subsection{Sensitivity to the Weights' Initialization}\label{Hyp_train}
This section evaluates the sensitivity of the proposed MS-NN-steer and the benchmark G-NN to the initialization of the learnable parameters during training. To perform this analysis, the initial guesses for the learnable parameters of the fully connected layers are randomized by sampling from a Gaussian distribution, with zero mean and a standard deviation of $10^{-4}$. For each network, $10$ different configurations are tested by varying the random seed used to initialize the weights. 

\begin{figure}[!h]
    \centering
    \includegraphics[width=\columnwidth]{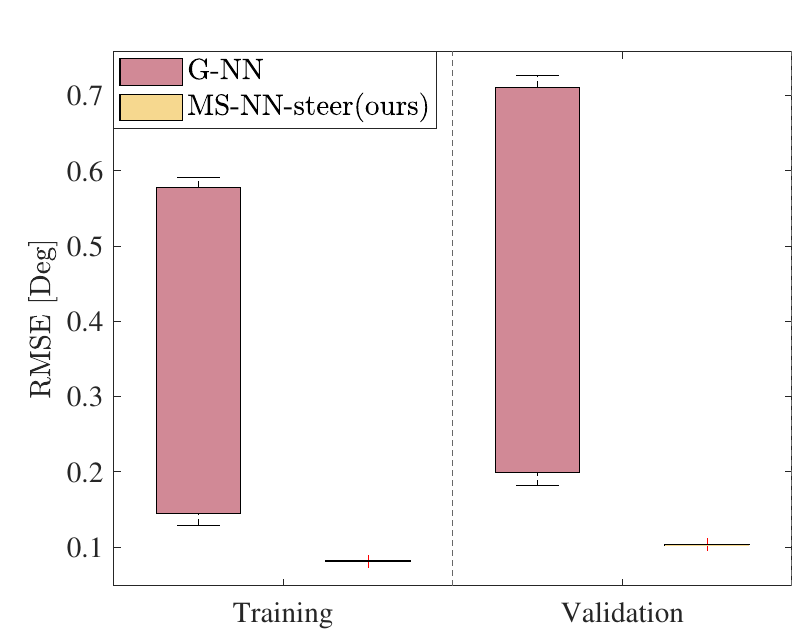}
    \caption{Boxplot of the RMSE values for the MS-NN-steer and G-NN, when trained with different weights' initialization. Our MS-NN-steer shows a very low variance in the RMSE values, suggesting that it is almost insensitive to the weights' initialization. In contrast, G-NN exhibits a high variance, indicating that its performance is sensitive to the choice of initial weights.}
    \label{fig:Seed}
\end{figure}

The results are shown in the boxplots of Fig. \ref{fig:Seed}. Our MS-NN-Steer is almost insensitive to the initialization of its weights, as indicated by the very small variance of the RMSE values across different seeds. This is a significant advantage, as it suggests that the model can achieve consistent performance regardless of the initial conditions. In contrast, the G-NN exhibits a very high variance in its RMSE values, indicating that its performance is more sensitive to the choice of initial weights, and thus it needs to be re-trained multiple times to achieve a satisfactory performance. 
 
Additionally, we conduct a related analysis by varying the learning rate in the range from $10^{-3}$ to $10^{-5}$. This investigation revealed a similar trend, confirming that our architecture maintains stable performance across a wide range of training conditions.

Hence, the proposed MS-NN-steer is robust to variations in its initial weights and learning rate, which is a desirable property since it reduces the need to perform extensive hyperparameter tuning and multiple training runs.

% The final analysis focuses on the influence of hyperparameter selection during the training phase. The goal of this investigation is to demonstrate that the performance of our proposed solution is relatively robust to variations in these hyperparameters. In particular, the model consistently achieves a comparable minimum RMSE across different training configurations, indicating a stable and reliable learning process. This behavior, however, does not hold for the generic network, which exhibits greater sensitivity to the choice of hyperparameters.

\section{Conclusion}
This paper presented a novel model-structured neural network (MS-NN-steer) for the steering control of a full-scale autonomous race car. The proposed architecture is based on the MS-NN framework, which incorporates physical priors and structured local models to improve the interpretability and generalization capabilities of the resulting neural network. We designed MS-NN-steer to capture the effect of the vehicle speed on the steering dynamics and handling diagram, and we showed that this leads to improved performance compared to the previous MS-NN-base architecture.

The proposed MS-NN-steer was trained and validated using real-world telemetry data from the TUM Autonomous Motorsport team, which won the A2RL competition in 2024. Our results demonstrate that the MS-NN-steer outperforms a general-purpose neural network, the A2RL baseline controller, and the previous MS-NN-base architecture. When trained with small datasets, our MS-NN-steer still generalizes well to unseen data, achieving a lower RMSE and FVU compared to the benchmarks. Finally, the MS-NN-steer's training is less sensitive to the initialization of the learnable parameters, which speeds up the training process and reduces the need for extensive hyperparameter tuning.

Future work will deploy the MS-NN-steer on the real car in the A2RL competition, and will test it in closed-loop when combined with trajectory planners \cite{Ogretmen2024,Piazza2024_mptree}.

\bibliography{Biblio.bib}
\bibliographystyle{IEEEtran}

\end{document}